# Edge Based Grid Super-Imposition for Crowd Emotion Recognition

Amol S Patwardhan

*Mechanical and Industrial Engineering Department, Louisiana State University*
apatwa3@lsu.edu

*Abstract—* **Numerous automatic continuous emotion detection system studies have examined mostly use of videos and images containing individual person expressing emotions. This study examines the detection of spontaneous emotions in a group and crowd settings. Edge detection was used with a grid of lines superimposition to extract the features. The feature movement in terms of movement from the reference point was used to track across sequences of images from the color channel. Additionally the video data capturing was done on spontaneous emotions invoked by watching sports events from group of participants. The method was view and occlusion independent and the results were not affected by presence of multiple people chaotically expressing various emotions. The edge thresholds of 0.2 and grid thresholds of 20 showed the best accuracy results. The overall accuracy of the group emotion classifier was 70.9%.**

## I. Introduction

Emotion recognition by computers has several applications in marketing and user profile specific advertisements. Many people watch live sports telecasts in groups. The football games in United States and international soccer games or basketball games are viewed by large crowds and groups of people gathering in a public place, sport bars and a friend's house. The various events during the games invoke several emotions among the supporters of each team. The emotions range from disgust, disappointment or joy depending on whether the supporter's team won or lost. There are incidents of controversial calls, surprise decisions and anguish when a referee makes a bad call or a favorite player gets injured and makes an excellent shot, which gets revoked. Multimodal emotion recognition has been studied using RGB-D data and audio-visual data. These studies have examined the videos mostly in controlled conditions using individual enactments or spontaneous emotional episodes.

In this research, the focus is on spontaneous group emotion detection under indoor lighting. Researchers [1] have examined the side effects of emotional thinking on memory and judgement. Emotion representation has been discussed in the field of psychology [2] using basic emotions such as happy, sad, angry, fear, surprise and disgust. In addition to video based emotion recognition, researchers [3] have also used vocal features to examine expression of emotions. A survey [4] on emotion detection and modelling using speech, audio and vocal input data has evaluated significance of such modalities in recognition accuracy. Application of emotion estimation in educational settings has been studied by researchers [5]. Studies [6] have used biosensors to assist in emotion detection and analyze the psychological and physiological effects of emotions in humans. Researchers [7] have used face, voice and body data to evaluate emotion modelling. A study [8] investigated the co-relation between various input channels in estimation accuracy using neural networks. Researchers [9] used dynamic Bayesian networks for monitoring crowd activity. Researchers [10, 11], [12] have studied intelligent surveillance systems using biosensors and bio-inspired devices. Studies [13] have evaluated the connection between body expressions and emotional states. Some studies [14], [15] have focussed on emotion maps and cognitive states and their relation with emotion modelling and expression. Studies [16], [17], [18], [19] have investigated dimensionality issues in emotion modelling, effect of cross-cultural influence of emotion, pedestrian behaviour and crowd simulation for emotion estimation.

Research [20], [21] has been done on human gait, pedestrian dynamics especially under influence of alcohol. Studies [22], [23] have focussed on specific emotion such as detection of fear, implementation strategies for automatic systems. Researchers have studied [24], [25] the view-invariant emotion detection for specific human behaviour such as crowd analysis, fatigue and tiredness and sleepiness prediction. In addition to audio-visual modalities studies [26], [27], [28] have focussed on text based sentiment analysis. Several supervised learning techniques (neural network, DBN, HMM) [29], [30], [31], [32], [33] for emotion recognition in various settings (closed spaces, indoor and outdoor) have been used to determine the accuracy of each method. Studies [34], [35], [36], [37], [38] on software implementation of automated continuous computer vision based multimodal emotion recognition techniques have been studied in detail. These studies have implemented novel methods to solve problem of modality fusion using hybrid methods and also provided architectural methods for real time detection. Researchers [39], [40], [41], [42], [43], [44], [45], [46], [47] have developed novel algorithms and behavioural rule based features for multimodal emotion recognition using continuous data and supervised learning.

## II. METHOD

Four groups of five people for 20 individuals participated in the study. The individuals were all dressed in casual attire and were aged in the range of 22 to 45. Eighteen participants were male and two were females. The groups gathered for different games of basketball finals. Three different cameras were used to record their reactions for 5 min at the beginning of each half and 5 min before the end of each half. The games usually get interesting at the beginning and towards the end. Tight games invoke more emotions compared to one-sided matches. The non-intrusive way of data capture allowed the subjects to watch the event and express their reactions without consciousness about being recorded on camera.

After the data was captured, the canny edge detection was applied on every third frame from the video sequence. This caused down sampling of a 24 frame per second to an eight frames per second sequence. For each image frame, the edge detection filter was applied. Then the frame was divided into 20 x 20 meshes and the intersection of the grid lines with the edges was considered as feature co-ordinates. For a consistent feature vector length, each line was further divided into 5 divisions and the features were counted as one for an intersection and zero for no intersection. Thus, 20 x 5 features for vertical lines and 20 x 5 features for horizontal lines on the mesh, for a total of 200 feature co-ordinates were obtained to form the feature vector. Additionally the temporal features were also tracked.

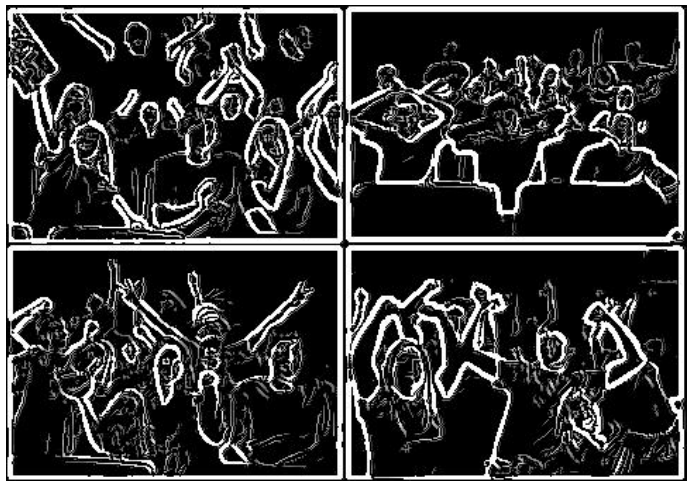

Fig. 1. Step 1 with edge detection applied to the frames

In the pre-processing step shown in Fig. 1. edge detection is applied to extract the series of high intensity change edges. These edges assisted in extracting further set of more discriminating feature points. This strategy was useful in extracting potential edges without depending on specific human behavioural actions, gestures and simply relied on view-independent image processing technique. Additionally, the occlusion caused by multiple people moving in the scene did not affect the process.

As the next step in the feature extraction process, the optimized grid was applied and super-imposed on each frame to find the intersections between the grid lines and the detected edges. Each intersection point was used as the feature for the frame. Next, the movement of each of these features was measured across the eight frames to find the temporal, kinetic and motion pattern of the feature. This allowed the method to rely on a limited yet discriminating set of features instead of tracking individual human activities. The method also allowed overcoming the limitation of wearing tracking devices to detect each human in the scene. It also allowed view-independent and occlusion resistance mechanism and purely relied on the visual layout of the scene based on the available data from the extracted features across various video frames and series of images. In addition, to limit the number of intersections between the edges and the lines of the grid only the N equally spaced points were considered. This number N was set to the same threshold as the number of blocks within the grid. For instance, if the grid size was 20 then the spacing was set to 20 as well. As a result, any other intersections were discarded. The settings for grid threshold at 5, 10, and 15 up to 50 showed that this strategy did not affect the overall accuracy results and the N simply contributed in limiting the feature vector size and eliminating the redundant tracked points.

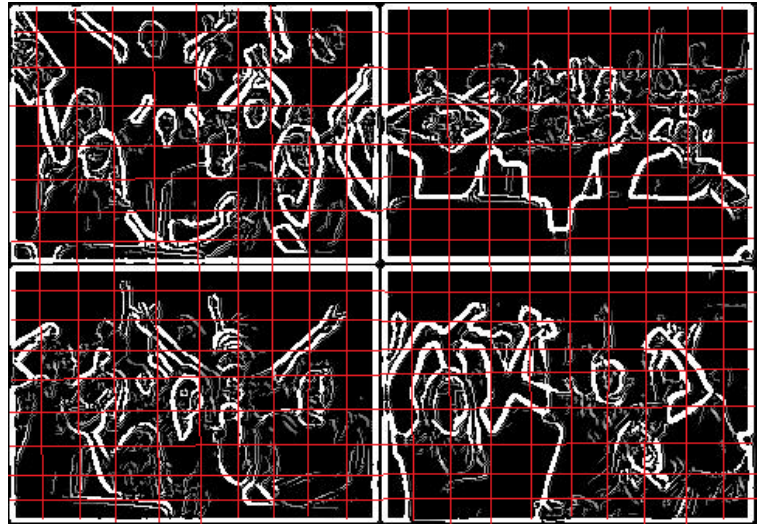

Fig. 2. Step 2 with the mesh applied to the edges.

The above figure shows the super-imposed grid and the intersection of the grid lines with the edges extracted from the first pre-processing step. The next figure shows the processing workflow to extract the edges, then the application of grid, mesh and extraction of potential points for tracking. This step is followed by tracking the motion of the points across the series of video frames.

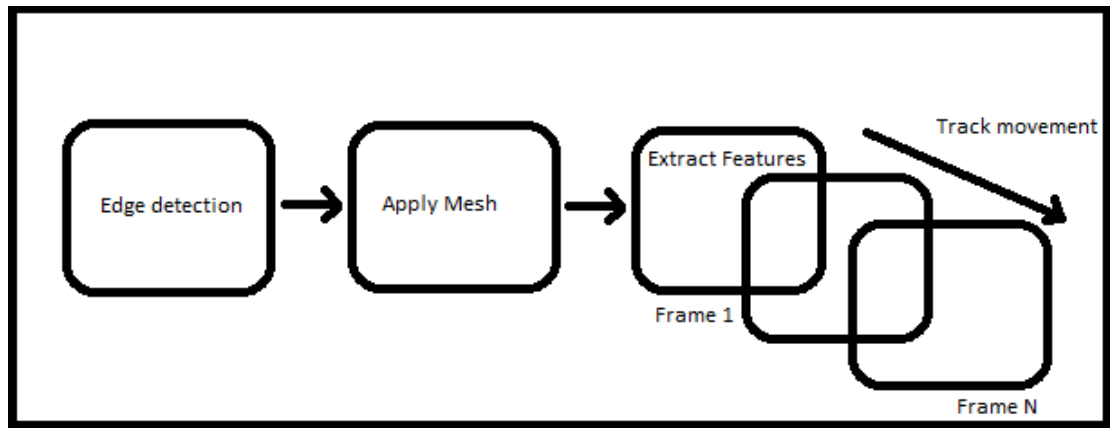

Fig. 3. Feature Extraction Process Workflow

The movement of each co-ordinate was tracked across the eight frames. Thus, the final feature vector consisted of 200 static and 200 velocity values across eight frames for 400 features. The best first search technique was used for feature selection. This resulted in reduced dimensionality of feature vector size and 23 discriminating features were chosen. The sequence of all actions was annotated using three observers to avoid inter-annotator disagreement.

The classes used were happy, angry, surprised, sad, disgust, fear and neutral. After the annotation was done, the feature vector was used to train the classifiers using support vector machine and radial basis function as the kernel function with 0.4 as the slack variable. The optimized slack variable was calculated using grid forward search method. The data was split into 70% training and 30% test data. The training was performed using 10-fold cross validation.

The optimal threshold for edge detection was obtained using classification tests on sample data set with grid size set to 10 through 50. Similarly, the optimal grid size was calculated using classification test and using six basic emotions and the neutral state as the candidate test emotions with the threshold set to 0.2 through 0.8.

III. RESULTS

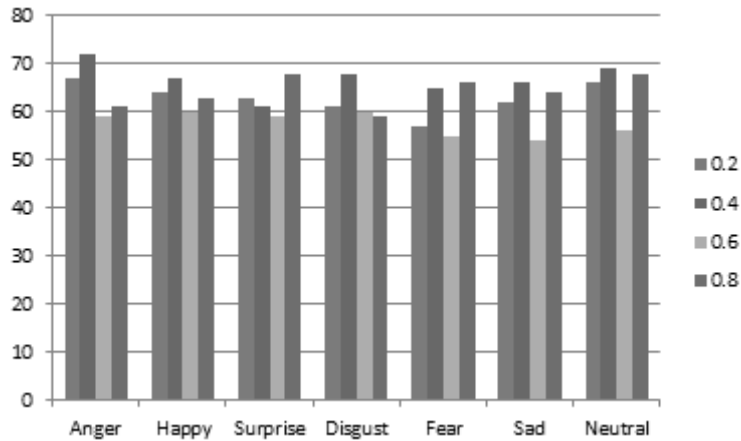

Fig. 4 Thresholds for edge detection

The threshold of 0.2 resulted in better accuracy for anger and neutral emotion. The threshold of 0.4 resulted in best accuracy for most of the emotions except fear and surprise. The thresholds for 0.6 and 0.8 performed poorly. This was because very few edges emerged after applying these thresholds resulting in fewer discriminating features in subsequent processing steps of grid based super-imposed feature extraction.

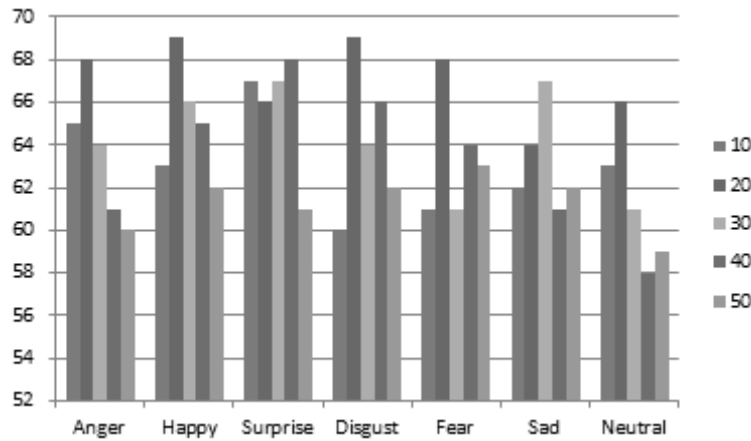

Fig. 5 Thresholds for grid size

| 0 | 1 | 2 | 3 | 4 | 5 | 6 | | |
|---|---|---|---|---|---|---|---|---|
| **0.678** | 0.036 | 0.046 | 0.039 | 0.081 | 0.072 | 0.048 | 0 | Anger |
| 0.099 | **0.766** | 0.025 | 0.05 | 0.022 | 0.005 | 0.032 | 1 | Happy |
| 0.109 | 0.023 | **0.72** | 0.037 | 0.062 | 0.031 | 0.018 | 2 | Surprise |
| 0.024 | 0.03 | 0.113 | **0.704** | 0.065 | 0.025 | 0.039 | 3 | Sad |
| 0.034 | 0.039 | 0.119 | 0.074 | **0.672** | 0.022 | 0.04 | 4 | Fear |
| 0.058 | 0.019 | 0.058 | 0.022 | 0.041 | **0.729** | 0.075 | 5 | Disgust |
| 0.085 | 0.04 | 0.068 | 0.027 | 0.065 | 0.025 | **0.691** | 6 | Neutral |

Fig. 6 Classification accuracy for six emotions

The grid size of 20 showed a clear improvement in accuracy as compared to the performance from grid size of 10 or 50. Except for surprise and sadness, the grid size of 20 showed the best emotion recognition accuracy. This was because the size was optimal. The size of 10 was too low to provide enough features while 50-grid size provided too many redundant and overlapping features to result in any significant improvement and at times even caused the accuracy to degrade.

The highest recall rate was observed for estimation of happy class label. It was mostly confused with surprise. The next best recall rates were recorded for surprise and disgust. The least accuracy was scene for anger. This was because in some cases the movement of individuals was so high in the scene, that they sometimes even walked out of the viewing frame or collapsed on the couch or hid their faces with objects such as gloves, helmets, caps or pillows. The above results were all recorded for the optimized edge detection thresholds of 0.4 and the optimized grid size threshold of 20 blocks per grid.

## IV. Conclusions

The threshold for edge detection that yielded the best recognition results was 0.4. The classification results for happiness emotion class label was the highest at 76.6% followed by disgust 72.9% and surprise 72%. The recognition rate for fear was the lowest with 67.2%. The overall accuracy of the group emotion recognition process was 70.9%. The grid size of 20 resulted in the best accuracy for 5 out of 7 emotion classes (including neutral class). This study evaluated a novel technique that implemented image-processing steps to extract the edges and then extract the features for chaotic scenes resulting from expression of emotions in crowded spectator settings. The group of people in the scene resulted in occluded view where many other techniques are not accurate because of the lack of patterns. As a result, the action based recognition techniques cannot be applied in such scenarios. The techniques mentioned in the paper showed promising results to overcome this limitation of view-dependence and lack of sufficient training data. This paper mostly looked into indoor scenes and a limited set of outdoor spontaneous scenes with crowds of people in the scene reacting emotionally expressive manner for sports events. As a future scope, the study could be extended in outdoor bigger crowd settings and a comparative study could be done between various techniques with the processing steps described in this paper.